\pdfoutput=1
\documentclass[11pt]{article}
\usepackage{EMNLP2023}
\usepackage{times}
\usepackage{latexsym}
\usepackage[T1]{fontenc}
\usepackage[utf8]{inputenc}
\usepackage{microtype}
\usepackage{inconsolata}
\usepackage{booktabs}

\usepackage[utf8]{inputenc} % allow utf-8 input
\usepackage[T1]{fontenc}    % use 8-bit T1 fonts
\usepackage{hyperref}       % hyperlinks
\usepackage{url}            % simple URL typesetting
\usepackage{booktabs}       % professional-quality tables
\usepackage{amsfonts}       % blackboard math symbols
\usepackage{nicefrac}       % compact symbols for 1/2, etc.
\usepackage{microtype}      % microtypography

\usepackage{tikz}
\usepackage{pgfplots}

\usepackage{enumitem}

\usepackage{xcolor}

\definecolor{darkgreen}{rgb}{0.0, 0.8, 0.0}
\definecolor{redgreen}{rgb}{0.8, 0.6, 0.0}
\definecolor{darkred}{rgb}{0.8, 0.0, 0.0}
\definecolor{lightred}{rgb}{0.95, 0.3, 0.4}
\definecolor{lightgreen}{rgb}{0.4, 0.7, 0.3}

\title{The Internal State of an LLM Knows When It's Lying}
\author{Amos Azaria\\
School of Computer Science,\\ Ariel University, Israel
\And
Tom Mitchell\\Machine Learning Dept.,\\ Carnegie Mellon University, Pittsburgh, PA}
\date{}
\pgfplotsset{compat=1.17}

\begin{document}

\maketitle

\begin{abstract}
    
While Large Language Models (LLMs) have shown exceptional performance in various tasks, one of their most prominent drawbacks is generating inaccurate or false information with a confident tone.  In this paper, we provide evidence that the LLM's internal state can be used to reveal the truthfulness of statements. This includes both statements provided to the LLM, and statements that the LLM itself generates.
Our approach is to train a classifier that outputs the probability that a statement is truthful, based on the hidden layer activations of the LLM as it reads or generates the statement.  Experiments demonstrate that given a set of test sentences, of which half are true and half false, our trained classifier achieves an average of 71\% to 83\% accuracy labeling which sentences are true versus false, depending on the LLM base model.   
Furthermore, we explore the relationship between our classifier's performance and approaches based on the probability assigned to the sentence by the LLM.  We show that while LLM-assigned sentence probability is related to sentence truthfulness, this probability is also dependent on sentence length and the frequencies of words in the sentence, resulting in our trained classifier providing a more reliable approach to detecting truthfulness, highlighting its potential to enhance the reliability of LLM-generated content and its practical applicability in real-world scenarios.
\end{abstract}

\section{Introduction}
Large Language Models (LLMs) have recently demonstrated remarkable success in a broad range of tasks \cite{brown2020language,bommarito2022gpt,driess2023palm,bubeck2023sparks}. However, when composing a response, LLMs tend to hallucinate facts and provide inaccurate information \cite{ji2023survey}. Furthermore, they seem to provide this incorrect information using confident and compelling language. The combination of a broad body of knowledge, along with the provision of confident but incorrect information, may cause significant harm, as people may accept the LLM as a knowledgeable source, and fall for its confident and compelling language, even when providing false information.

%why does it hallucinate facts?
We believe that in order to perform well, an LLM must have some internal notion as to whether a sentence is true or false, as this information is required for generating (or predicting) following tokens. For example, consider an LLM generating the following false information ``The sun orbits the Earth." After stating this incorrect fact, the LLM is more likely to attempt to correct itself by saying that this is a misconception from the past. But after stating a true fact, for example ``The Earth orbits the sun," it is more likely to focus on other planets that orbit the sun. Therefore, we hypothesize that the truth or falsehood of a statement should be represented by, and therefore extractable from, the LLM's internal state. 

\begin{figure}
  \centering
  \resizebox{\columnwidth}{!}{
\begin{tikzpicture}[scale=1.4, level distance=2cm,
    level 1/.style={sibling distance=2cm},
    level 2/.style={sibling distance=3cm},
    every node/.style={circle, draw, text width=2cm, align=center},
    edge from parent/.style={draw,-latex}]
  % draw the root node
  \node[fill=green!20!white] {Pluto is the}
    % add the "second" node
    child {node[fill=gray!20!white] {second}}
    % add the "largest" node
    child {node[fill=gray!20!white] {largest}}
    % add the "smallest" node
    child {node[fill=green!20!white] {smallest}
      % add the "dwarf planet in our solar system" node
      child {node[fill=green!20!white] {dwarf planet in our solar system.}}
      % add the "celestial body in the solar system that has ever been classified as a planet" node
      child {node[fill=red!20!white] {celestial body in the solar system that has ever been classified as a planet.}}}
    % add the "only" node
    child {node[fill=gray!20!white] {only}}
    % add the "most" node
    child {node[fill=gray!20!white] {most}};
\end{tikzpicture}
}
\caption{A tree diagram that demonstrates how generating words one at a time and committing to them may result in generating inaccurate information.}
\label{fig:pluto-tree}
\end{figure}
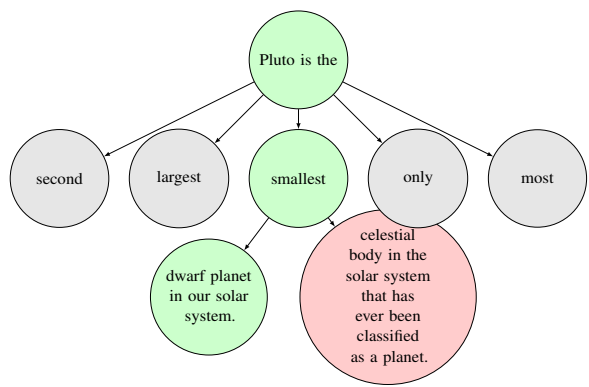

Interestingly, retrospectively ``understanding'' that a statement that an LLM has just generated is false does not entail that the LLM will not generate it in the first place. We identify three reasons for such behavior. The first reason is that an LLM generates a token at a time, and it ``commits'' to each token generated. Therefore, even if maximizing the likelihood of each token given the previous tokens, the overall likelihood of the complete statement may be low. For example, consider a statement about Pluto. The statement begins with the common words "Pluto is the", then, since Pluto used to be the smallest planet the word "smallest" may be a very plausible choice. Once the sentence is "Pluto is the smallest", completing it correctly is very challenging, and when prompted to complete the sentence, GPT-4 (March 23rd version) completes it incorrectly: ``Pluto is the smallest dwarf planet in our solar system.'' In fact, Pluto is the second largest dwarf planet in our solar system (after Eris). One plausible completion of the sentence correctly is ``Pluto is the smallest celestial body in the solar system that has ever been classified as a planet.'' (see Figure \ref{fig:pluto-tree}). %The following prompt also resulted in a mistake by GPT-4: Complete the following sentence using an adjective: Pluto is the ... Pluto is the farthest (or most distant) planet (dwarf planet) from the sun in our solar system.
%\textcolor{blue}{To Tom: I'm not sure if the example I used is what you were looking for, though I actually really like it. 
Consider the following additional example: ``Tiztoutine is a town in Africa located in the republic of Niger.'' Indeed, Tiztoutine is a town in Africa, and many countries' names in Africa begin with ``the republic of''. However, Tiztoutine  is located in Morocco, which is not a republic, so once the LLM commits to ``the republic of'', it cannot complete the sentence using ``Morocco", but completes it with ``Niger". In addition, committing to a word at a time may lead the LLM to be required to complete a sentence that it simply does not know how to complete. For example, when describing a city, it may predict that the next words should describe the city's population. Therefore, it may include in a sentence "Tiztoutine has a population of", but the population size is not present in the dataset, so it must complete the sentence with a pure guess. 
%Prompt to GPT-4:Add a single word to the following sentence: "Tiztoutine is a town in Africa located in the republic of" GPT-4 response: Tiztoutine is a town in Africa located in the republic of Niger.

The second reason for an LLM to provide false information is that at times, there may be many ways to complete a sentence correctly, but fewer ways to complete it incorrectly. Therefore, it might be that a single incorrect completion may have a higher likelihood than any of the correct completions (when considered separately). 

Finally, since it is common for an LLM to not use the maximal probability for the next word, but to sample according to the distribution over the words, it may sample words that result in false information.

In this paper we present our Statement Accuracy Prediction, based on Language Model Activations (SAPLMA). SAPLMA is a simple yet powerful method for detecting whether a statement generated by an LLM is truthful or not. Namely, we build a classifier that receives as input the activation values of the hidden layers of an LLM. The classifier determines for each statement generated by the LLM if it is true or false. Importantly, the classifier is trained on out-of-distribution data, which allows us to focus specifically on whether the LLM has an internal representation of a statement being true or false, regardless of the statement's topic. 

%The data is not used, as done in classical machine learning approaches, to determine which statements are true or false,
In order to train SAPLMA we created a dataset of true and false statements from 6 different topics. Each statement is fed to the LLM, and its hidden layers' values are recorded. The classifier is then trained to predict whether a statement is true or false only based on the hidden layer's values. Importantly, our classifier is not tested on the topics it is trained, but on a held-out topic. We believe that this is an important measure, as it requires SAPLMA to extract the LLM's internal belief, rather than learning how information must be aligned to be classified as true.  
We show that SAPLMA, which leverages the LLM's internal states, results in a better performance than prompting the LLM to explicitly state whether a statement is true or false. Specifically, SAPLMA reaches accuracy levels of between 60\% to 80\% on specific topics, while few-shot prompting achieves only slightly above random performance, with no more than a 56\% accuracy level. 

Of course there will be some relationship between the truth/falsehood of a sentence, and the probability assigned to that sentence by a well-trained LLM.  But the probability assigned by an LLM to a given statement depends heavily on the frequency of the tokens in the statement as well as its length. Therefore, sentence probabilities provide only a weak signal of the truth/falsehood of the sentence.  At minimum, they must be normalized to become a useful signal of the veracity of a statement. As we discuss later, SAPLMA classifications of truth/falsehood significantly outperform simple sentence probability.  In one test described later in more detail, we show that SAPLMA performs well on a set of LLM-generated sentences that contain 50\% true, and 50\% false statements.  We further discuss the relationship between statement probability and veracity in the discussion section.

SAPLMA employs a simple and relatively shallow feedforward neural network as its classifier, which requires very little computational power at inference time. Therefore, it can be computed alongside the LLM's output. We propose for SAPLMA to supplement an LLM presenting information to users. If SAPLMA detects that the LLM ``believes'' that a statement that it has just generated is false, the LLM can mark it as such. %This is in contrast to current LLMs that cannot admit making a mistake during the generation of a response, and must continue their response in a way consistent to all information that has already been generated.
This could raise human trust in the LLM responses. Alternatively, the LLM may merely delete the incorrect statement and generate a new one instead.

To summarize, the contribution of this paper is twofold. 
\begin{itemize}[topsep=0pt,itemsep=-1ex,partopsep=1ex,parsep=1ex]
    \item The release of a dataset of true-false statements along with a method for generating such information. 
    \item Demonstrating that an LLM might ``know'' when a statement that it has just generated is false, and proposing SAPLMA, a method for extracting this information.
\end{itemize}

\section{Related Work}
In this section we provide an overview of previous research on LLM hallucination, accuracy, and methods for detecting false information, and we discuss datasets used to that end.

Many works have focused on hallucination in machine translation \cite{dale2022detecting,ji2023survey}. For example, Dale et al. \cite{dale2022detecting} consider hallucinations as translations that are detached from the source, hence they propose a method that evaluates the percentage of the source contribution to a generated translation.
If this contribution is low, they assume that the translation is detached from the source and is thus considered to be hallucinated. Their method improves detection accuracy hallucinations. The authors also propose to use multilingual embeddings, and compare the similarity between the embeddings of the source sentence and the target sentence. If this similarity is low, the target sentence is considered to be hallucinated. The authors show that the latter method works better. However, their approach is very different than ours, %the work presented in this paper, 
as we do not assume any pair of source and target sentences. In addition, while we also use the internal states of the model, we do so by using the hidden states to descriminate between statements that the LLM ``believes'' are true and those that are false. Furtheremore, we focus on detecting false statements rather than hallucination, as defined by their work. 

Other works have focused on hallucination in text summarization \cite{pagnoni2021understanding}. Pagnoni et al. propose a benchmark for factuality metrics of text summarization. Their benchmark is developed by gathering summaries from several summarization methods and requested humans to annotate their errors. The authors analyze these anotation and the proportion of the different factual error of the summarization methods.
We note that most works that consider hallucination do so with relation to a given input (e.g., a passage) that the model operates on. For example, a summarization model that outputs information that does not appear in the provided article, is considered hallucinate it, regardless if the information is factually true. However, in this work we consider a different problem----the veracity of the output of an LLM, without respect to a specific input. % is providing incorrect information without assuming any specific input or task.

% there have been several attempts...
Some methods for reducing hallucination assume that the LLM is a black box \cite{peng2023check}. This approach uses different methods for prompting the LLM, possibly by posting multiple queries for achieving better performance. 
Some methods that can be used for detecting false statements may include repeated queries and measuring the discrepancy among them. We note that instead of asking the LLM to answer the same query multiple times, it is possible to request the LLM to rephrase the query (without changing its meaning or any possible answer) and then asking it to answer each rephrased question.

Other methods finetune the LLM, using human feedback, reinforcement learning, or both \cite{bakker2022fine,ouyang2022training}. Ouyang et al. propose a method to improve %the factual correctness of 
LLM-generated content using reinforcement learning from human feedback. Their approach focuses on fine-tuning the LLM with a reward model based on human judgments, aiming to encourage the generation of better content. %more factually correct outputs.
However, fine tuning, in general, may cause a model to not perform as well on other tasks \cite{kirkpatrick2017overcoming}. In this paper, we take an intermediate approach, that is, we assume access to the model parameters, but do not fine-tune or modify them. Another approach that can be applied to our settings is presented by \cite{burns2022discovering}, named Contrast-Consistent Search (CCS). However, CCS requires rephrasing a statement into a question, evaluating the LLM on two different version of the prompt, and requires training data from the same dataset (topic) as the test set. These limitations render it unsuitable for running in practice on statements generated by an LLM. In addition, CCS increases the accuracy by only approximately 4\% over the 0-shot LLM query, while our approach demonstrates a nearly 20\% increase over the 0-shot LLM

A dataset commonly used for training and fine-tuning LLMs is the Wizard-of-Wikipedia \cite{dinan2018wizard}. The Wizard-of-Wikipedia dataset includes interactions between a human apprentice and a human wizard. The human wizard receives relevant Wikipedia articles, which should be used to select a relevant sentence and compose the response. The goal is to replace the wizard with a learned agent (such as an LLM). Another highly relevant dataset is FEVER \cite{thorne2018fever,thorne2019fever2}. The FEVER dataset is designed for developing models that receive as input a claim and a passage, and must determine whether the passage supports the claim, refutes it, or does not provide enough information to support or refute it. While the FEVER dataset is highly relevant, it does not provide simple sentence that are clearly true or false independently of a provided passage. In addition, the FEVER dataset is not partitioned into different topics as the true-false dataset provided in this paper.

In conclusion, while several approaches have been proposed to address the problem of hallucination and inaccuracy in automatically generated content, our work is unique in its focus on utilizing the LLM's hidden layer activations to determine the veracity of generated statements. Our method offers the potential for more general applicability in real-world scenarios, operating alongside an LLM, without the need for fine-tuning or task-specific modifications.

\section{The True-False Dataset}
The work presented in this paper requires a dataset of true and false statements. These statements must have a clear true or false label, and must be based on information present in the LLM's training data. Furthermore, since our approach intends to reveal that the hidden states of an LLM have a notion of a statement being true or false, the dataset must cover several disjoint topics, such that a classifier can be trained on the LLM's activations of some topics while being tested on another. Unfortunately, we could not find any such dataset and therefore, compose the true-false dataset. 

Our true-false dataset covers the following topics: ``Cities", ``Inventions", ``Chemical Elements", ``Animals", ``Companies", and ``Scientific Facts". For the first 5 topics, we used the following method to compose the dataset. We used a reliable source\footnote{Cities: Downloaded from simplemaps.
Inventions: Obtained from Wikipedia list of inventors.
Chemical Elements: Downloaded from pubchem ncbi nlm nih gov periodic-table.
Animals: Obtained from kids national geographics.
Companies: Forbes Global 2000 List 2022: The Top 200.
} that included a table with several properties for each instance. For example, for the ``chemical elements" we used a table that included, for each element, its name, atomic number, symbol, standard state, group block, and a unique property (e.g., Hydrogen, 1, H, Gas, Nonmetal, the most abundant element in the universe). For each element we composed true statement using the element name and one of its properties (e.g., ``The atomic number of Hydrogen is 1''). Then, we randomly selected a different row for composing a false statement (e.g., ``The atomic number of Hydrogen is 34''). If the value in the different row is identical to the value in the current row, we resample a different row until we obtain a value that is different.
This process was repeated for the all topics except the ``Scientific Facts''. 
For the ``Scientific Facts'' topic, we asked ChatGPT (Feb 13 version) to provide ``scientific facts that are well known to humans'' (e.g. ``The sky is often cloudy when it's going to rain''). We then asked ChatGPT to provide the opposite of each statement such that it becomes a false statement (e.g., ``The sky is often clear when it's going to rain''). The statements provided by ChatGPT were manually curated, and verified by two human annotators. The classification of 48 facts were questioned by at least one of the annotators; these facts were removed from the dataset. %interestingly, only one correct fact was questioned by one of the annotators, while 47 of the incorrect facts were question by at least one of the annotators.
%As a result of the curation method, the dataset maintains a well-balanced distribution of true and false statements.
%Table \ref{tbl:datasets} presents the number of statements in each topic.
The true-false dataset comprises 6,084 sentences, including 1,458 sentences for ``Cities", 876 for ``Inventions", 930 for ``Chemical Elements", 1,008 for ``Animals", 1,200 for ``Companies", and 612 for ``Scientific Facts".
The following are some examples of \emph{true} statements from the dataset:
\begin{itemize}[topsep=0pt,itemsep=-1ex,partopsep=1ex,parsep=1ex]
    \item Cities: ``Oranjestad is a city in Aruba''
    \item Inventions: ``Grace Hopper invented the COBOL programming language''
    %\item Chemical Elements: ``Boron is used in the production of glass and ceramics''
    \item Animals: ``The llama has a diet of herbivore''
    \item Companies: ``Meta Platforms has headquarters in United States''
    \item Scientific Facts: ``The Earth's tides are primarily caused by the gravitational pull of the moon''
\end{itemize}
The following are some examples of \emph{false} statements from the dataset:
\begin{itemize}[topsep=0pt,itemsep=-1ex,partopsep=1ex,parsep=1ex]
    %\item Cities: ``Lilongwe is a city in Saint Lucia'' %``Wellington is a name of a country''
    %\item Inventions: ``David Schwarz lived in France''
    \item Chemical Elements: ``Indium is in the Lanthanide group''
    \item Animals: ``The whale has a long, tubular snout, large ears, and a powerful digging ability to locate and consume termites and ants.''
    %\item Companies: ``KDDI operates in the industry of Materials''
    \item Scientific Facts: ``Ice sinks in water due to its higher density''
\end{itemize}
Other candidates for topics to be added to the dataset include sports, celebrities, and movies. The true-false dataset is available at: \url{azariaa.com/Content/Datasets/true-false-dataset.zip}.

%Amos: removed to save space:
% \begin{table}[ht]
% \centering
% \begin{tabular}{l r}
% \toprule
% Topic & Number of Rows \\
% \midrule
% Cities & 1458 \\
% Inventions & 876 \\
% Chemical Elements & 930 \\
% Animals & 1008 \\
% Companies & 1200 \\
% Scientific Facts & 612\\
% \midrule
% Total & 6,084 \\
% \bottomrule
% \end{tabular}
% \caption{Number of rows for each topic in the true-false dataset.}
% \label{tbl:datasets}
% \end{table}

%% Using Inner Layer Value Activations Predicting Statement Truthfulness (Accuracy)
% PANDA
% SAPLMA Statement Accuracy Prediction, based on Language Model Activations
\section{SAPLMA} %Methodology}
\label{sec:SAPLMA}
In this section, we present our Statement Accuracy Prediction, based on Language Model Activations (SAPLMA), a method designed to determine the truthfulness of statements generated by an LLM using the values in its hidden layers. 
Our general hypothesis is that the values in the hidden layers of an LLM contain information on whether the LLM ``believes'' that a statement is true or false. However, it is unclear which layer should be the best candidate for retaining such information. %While the last hidden layer should retain such information, it is most likely to focus on generating the next token; however layers that are too close to the input are likely focused on extracting lower-level information from the input. 
While the last hidden layer should contain such information, it is primarily focused on generating the next token. Conversely, layers closer to the input are likely focused on extracting lower-level information from the input.
Therefore, we use several hidden layers as candidates. We use two different LLMs:  Facebook OPT-6.7b \cite{zhang2022opt} and LLAMA2-7b \cite{roumeliotis2023llama}; both composed of 32 layers. For each LLM, we compose five different models, each using activations from a different layer. Namely, we use the last hidden layer, the 28th layer (which is the 4th before last), the 24th layer (which is the 8th before last), the 20th layer (which is the 12th before last), and the middle layer (which is the 16th layer). We note that each layer is composed of 4096 hidden units.

SAPLMA's classifier employs a feedforward neural network, featuring three hidden layers with decreasing numbers of hidden units (256, 128, 64), all utilizing ReLU activations. The final layer is a sigmoid output. We use the Adam optimizer. We do not fine-tune any of these hyper-parameters for this task. The classifier is trained for 5 epochs. 

For each topic in the true-false dataset, we train the classifier using only the activation values obtained from all other topics and test its accuracy on the current topic. This way, the classifier is required to determine which sentences the LLM ``believes'' are true and which it ``believes'' are false, in a general setting, and not specifically with respect to the topic being tested. To obtain more reliable results, we train each classifier three times with different initial random weights. This process is repeated for each topic, and we report the accuracy mean over these three runs.

\section{Results}

We compare the performance of SAPLMA against three different baselines. The first is BERT, for which we train a classifier (with an identical architecture to the one used by SAPLMA) on the BERT embeddings of each  sentence. %Since we train on the embeddings of all topics except the topic that is tested, we expect that the BERT embeddings will perform very poorly, as they focus on the semantics of the statement rather than on whether a statement is true or false. 
Our second baseline is a few shot-learner using OPT-6.7b. This baseline is an attempt to reveal whether the LLM itself ``knows'' whether a statement is true or false. Unfortunately, any attempts to explicitly prompt the LLM in a `zero-shot'' manner to determine whether a statement is true or false completely failed with accuracy levels not going beyond 52\%. Therefore, we use a few shot-query instead, which provided the LLM with truth values from the same topic it is being tested on. Note that this is very different from our methodology, but was necessary for obtaining some results. Namely, we provided few statements along with the ground-truth label, and then the statement in question. We recorded the probability that the LLM assigned to the token ``true'' and to the token ``false''. Unfortunately, the LLM had a tendency to assign higher values to the ``true'' token; therefore, we divided the probability assigned to the ``true'' token by the one assigned to the ``false'' token. Finally, we considered the LLM's prediction ``true'' if the value was greater than the average, and ``false'' otherwise. We tested a 3-shot and a 5-shot version. The third baseline, given a statement `X', we measure the probabilities (using OPT-6.7) of the sentences “It is true that X”, and “It is false that X”, and pick the higher probability (considering only the probabilities for X, not the added words). This normalizes the length and frequency factors. 

\begin{table}[ht]
\centering
\resizebox{\columnwidth}{!}{
\begin{tabular}{l|rrrrrr|r}
\toprule
\textbf{Model} & \textbf{Cities} & \textbf{Invent.} & \textbf{Elements} & \textbf{Animals} & \textbf{Comp.} & \textbf{Facts} & \textbf{Average} \\
\midrule
last-layer & 0.7796 & 0.5696 & 0.5760 & 0.6022 & 0.6925 & 0.6498 & 0.6449 \\
28th-layer & 0.7732 & 0.5761 & 0.5907 & 0.5777 & 0.7247 & 0.6618 & 0.6507\\
% 27th-layer & 0.8025 & 0.6005 & 0.5924 & 0.5942 & 0.7314 & 0.6477 & 0.6614 \\
24th-layer & 0.7963 & 0.6712 & \textbf{0.6211} & 0.5800 & 0.7758 & \textbf{0.6868} & 0.6886 \\
20th-layer & \textbf{0.8125} & \textbf{0.7268} & 0.6197 & \textbf{0.6058} & \textbf{0.8122} & 0.6819 & \textbf{0.7098} \\
middle-layer & 0.7435 & 0.6400 & 0.5645 & 0.5800 & 0.7570 & 0.6237 & 0.6515 \\
\midrule
BERT & 0.5357 & 0.5537 & 0.5645 & 0.5228 & 0.5533 & 0.5302 & 0.5434 \\
3-shot & 0.5410 & 0.4799 & 0.5685 & 0.5650 & 0.5538 & 0.5164 & 0.5374 \\
5-shot & 0.5416 & 0.4799 & 0.5676 & 0.5643 & 0.5540 & 0.5148 & 0.5370 \\
It-is-true&0.523&0.5068&0.5688&0.4851&0.6883&0.584& 0.5593 \\
\bottomrule
\end{tabular}
}
\caption{Accuracy classifying truthfulness of externally generated sentences during reading. The table shows accuracy of all the models tested for each of the topics, and average accuracy using OPT-6.7b as the LLM.}
\label{tbl:results}
\end{table}

% \begin{table}[h]
% \centering
% \begin{tabular}{l c c }
% \toprule
% \textbf{Model} & \textbf{Average Threshold} & \textbf{Average Accuracy} \\ \midrule
% last-layer & 0.5144 & 0.6860 \\ 
% 28th-layer & 0.5094 & 0.7021 \\ 
% 24th-layer & 0.5249 & 0.7251  \\ 
% 20th-layer & 0.5624 & 0.7251  \\ 
% middle-layer & 0.5350 & 0.6741 \\
% \midrule
% BERT & 0.5127 & 0.5428\\
% \bottomrule
% \end{tabular}
% \caption{Averages of average threshold and average accuracy for each of the models when using 30\% of the test data to determine the optimal threshold.}
% \label{table:averages}
% \end{table}

\begin{figure}
    \centering
    \resizebox{\columnwidth}{!}{
\begin{tikzpicture}
    \begin{axis}[
        ybar,
        bar width=0.25cm,
        width=\textwidth,
        height=.5\textwidth,
        ylabel={Accuracy},
        ymin=0.5,
        ymax=1,
        symbolic x coords={Cities, Invent., Elements, Animals, Comp., Facts, Average},
        xtick=data,
        xtick style={draw=none},
        x tick label style={rotate=45, anchor=east},
        legend style={at={(0.5,-0.25)}, anchor=north, legend columns=-1},
        enlarge x limits=0.1
        ]
        
        % SAPLMA 20th-layer
        \addplot coordinates {
            (Cities, 0.8125)
            (Invent., 0.7268)
            (Elements, 0.6197)
            (Animals, 0.6058)
            (Comp., 0.8122)
            (Facts, 0.6819)
            (Average, 0.7098)
        };

        % It-is-true
        \addplot coordinates {
            (Cities, 0.523)
            (Invent., 0.5068)
            (Elements, 0.5688)
            (Animals, 0.4851)
            (Comp., 0.6883)
            (Facts, 0.584)
            (Average, 0.5593)
        };
        
        % BERT
        \addplot coordinates {
            (Cities, 0.5357)
            (Invent., 0.5537)
            (Elements, 0.5645)
            (Animals, 0.5228)
            (Comp., 0.5533)
            (Facts, 0.5302)
            (Average, 0.5434)
        };

        % 3-shot
        \addplot coordinates {
            (Cities, 0.5410)
            (Invent., 0.4799)
            (Elements, 0.5685)
            (Animals, 0.5650)
            (Comp., 0.5538)
            (Facts, 0.5164)
            (Average, 0.5374)
        };
        
        % 5-shot
        \addplot coordinates {
            (Cities, 0.5416)
            (Invent., 0.4799)
            (Elements, 0.5676)
            (Animals, 0.5643)
            (Comp., 0.5540)
            (Facts, 0.5148)
            (Average, 0.5370)
        };

        \legend{SAPLMA, It-is-true, BERT, 3-shot, 5-shot}
    \end{axis}
\end{tikzpicture}
}
\caption{A bar-chart comparing the accuracy of SAPLMA (20th-layer), BERT, 3-shot, 5-shot, and It-is-true on the 6 topics, and the average. SAPLMA consistently outperforms other models across all categories. Since the data is balanced, a random classifier should obtain an accuracy of 0.5.  }
\label{fig:results}
\end{figure}
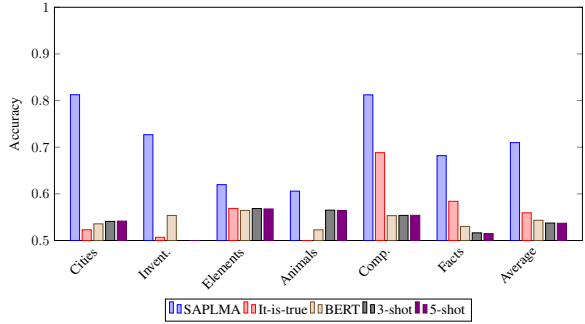

Table \ref{tbl:results} and Figure \ref{fig:results} present the accuracy of all the models tested using the OPT-6.7b LLM, for each of the topics, along with the average accuracy. As depicted by the table and figure, SAPLMA clearly outperforms BERT and Few-shot learning, with BERT, 3-shot, and 5-shot learning achieving only slightly above a random guess ($0.50$).
It can also be observed that SAPLMA for OPT-6.7b seems to work best when using the 20th layer (out of 32). Recall that each model was trained three times. We note that the standard deviation between the accuracy among the different runs was very small; therefore, the differences in performance between the different layers seem consistent. However, the optimal layer to be used for SAPLMA is very likely to depend on the LLM. %, but obtaining the activations from the third quartile seems to perform well. 

We note that the average \emph{training} accuracy for SAPLMA (using OPT-6.7b's 20th layer) is 86.4\%. We believe that this relatively high value may indicate once again that the veracity of a statement is inherit to the LLM. To demonstrate this, we run a test where the true/false labels are randomly permuted. The average accuracy on the random training data is only 62.5\%. This indicates that our model does not have the capacity to completely over-fit the training data, and thus, must exploit structure and patterns that appear in the data.

\begin{table}[ht]
\centering
\resizebox{\columnwidth}{!}{
\begin{tabular}{l|rrrrrr|r}
\toprule
\textbf{Model} & \textbf{Cities} & \textbf{Invent.} & \textbf{Elements} & \textbf{Animals} & \textbf{Comp.} & \textbf{Facts} & \textbf{Average} \\
\midrule
last-layer & 0.7574 & 0.6735 & 0.6814 & 0.7338 & 0.6736 & 0.7444 & 0.7107 \\
28th-layer & 0.8146 & 0.7207 & 0.6767 & 0.7249 & 0.6894 & 0.7662 & 0.7321 \\
24th-layer & 0.8722 & 0.7816 & 0.6849 & 0.7394 & 0.7094 & 0.7858 & 0.7622 \\
20th-layer & 0.8820 & 0.8459 & 0.6950 & 0.7758 & 0.8319 & 0.8053 & 0.8060 \\
16th-layer & \textbf{0.9223} & \textbf{0.8938} & \textbf{0.6939} & \textbf{0.7774} & \textbf{0.8658} & \textbf{0.8254} & \textbf{0.8298} \\
\bottomrule
\end{tabular}
}
\caption{Accuracy classifying truthfulness of externally generated sentences using SAPLMA with LLAMA2-7b. The table shows accuracy of all the models tested for each of the topics, and the average accuracy.}
\label{tbl:llama2results}
\end{table}

Additionally, Table \ref{tbl:llama2results} presents the performance of SAPLMA using LLAMA2-7b. As expected, SAPLMA using LLAMA2-7b achieves much better performance. Interestingly, for LLAMA2-7b, the middle layer performs best.

As for the differences between the topics, we believe that these values depend very much on the training data of the LLM. That is, we believe that the data used for training OPT-6.7b and LLAMA2-7b includes information or stories about many cities and companies, and not as much on chemical elements and animals (other than the very common ones). Therefore, we conjecture that is the reason SAPLMA achieves high accuracy for the ``cities'' topic (while trained on all the rest) and the ``companies'' topic, but achieves much lower accuracy when tested on the ``animals'' and ``elements'' topics.
%We note that the variance between the three ... was only ...

%Figure \ref{fig:res} depicts the results of the

In addition to the topics from the true-false dataset, we also created a second data set of statements generated by the LLM itself (the OPT-6.7b model). For generating statements, we provided a true statement not present in the dataset, and allowed the LLM to generate a following statement.
We first filtered out any statements that were not factual statements (e.g., ``I'm not familiar with the Japanese versions of the games.''). All statements were generated using the most probable next word at each step, i.e., we did not use sampling.
This resulted in $245$ labeled statements. The statements were fact-checked and manually labeled by three human judges based on web-searches. The human judges had a very high average observed agreement of 97.82\%, and an average Cohen's Kappa of 0.9566. The majority determined the ground-truth label for each statement. 48.6\% of the statements were labeled as true, resulting in a balanced dataset.

Each of the models was trained $14$ times using the same classifier described in Section \ref{sec:SAPLMA}. The models were trained
on the entire true-false dataset (i.e., all topics, but not the generated sentences) and tested on the generated sentences.

\begin{table}[ht]
\centering
\begin{tabular}{l r r}
\toprule
\textbf{Model} & \textbf{Accuracy} & \textbf{AUC}\\
\midrule
last-layer & 0.6187 & 0.7587\\
28th-layer & \textbf{0.6362} & \textbf{0.7614}\\
24th-layer & 0.6134 & 0.7435\\
20th-layer & 0.6029 & 0.7182\\
middle-layer & 0.5566 & 0.6610\\
\midrule
BERT & 0.5115 & 0.5989 \\
3-shot & 0.5041 & 0.4845 \\
5-shot & 0.5125 & 0.4822\\
\bottomrule
\end{tabular}
\caption{Accuracy classifying truthfulness of sentences generated by the LLM (OPT-6.7b) itself.}
\label{tbl:generated}
\end{table}

Table \ref{tbl:generated} presents the average accuracy of all models on the sentences generated by the LLM.
As anticipated, SAPLMA (using OPT-6.7b) clearly outperforms the baselines, which appear to be entirely ineffectual in this task, achieving an accuracy near 50\%.
However, the accuracy of SAPLMA on these sentences is not as promising as the accuracy achieved when tested on some of the topics in the true-false dataset (i.e., the cities and companies). Since we expected the LLM to generate sentences that are more aligned with the data it was trained on, we did expect SAPLMA's performance on the generated sentences to be closer to its performance on topics such as cities and companies, which are likely aligned with the data the LLM was trained on. However, there may be also a counter-effect in play: the sentences in the true-false dataset were mostly generated using a specific pattern (except the scientific facts topic), such that each sentence is clearly either true or false. However, the sentences generated by the LLM where much more open, and their truth value may be less clearly defined (despite being agreed upon by the human judges). For example, one of the sentences classified by all human judges as false is ``Lima gets an average of 1 hour of sunshine per day.'' However, this sentence is true during the winter. Another example is ``Brink is a river,'' which was also classified as false by all three human judges; however, brink refers to river bank (but is not a name of a specific river, and does not mean river).
Indeed, SAPLMA classified approximately 70\% of the sentences as true, and the AUC values seem more promising. This may hint that any sentence that seems plausible is classified as true. Therefore, we evaluate the models  using 30\% of the generated sentences for determining which threshold to use, i.e., any prediction above the threshold is considered true. Importantly, we do not use this validation set for any other goal. We test the models on the remaining 70\% of the generated sentences.
%set the positive threshold at 0.8. That is, a prediction is considered true if the probability returned by the model is greater than 0.8. 
We do not evaluate the few shot models again, as our evaluation guaranteed that the number of positive predictions would match the number of negative predictions, which matches the distribution in of the data.

\begin{table}[ht]
\centering
\begin{tabular}{l r r}
\toprule
\textbf{Model} & \textbf{Avg Threshold} & \textbf{Accuracy}\\
\midrule
last-layer &  0.8687 & 0.7052 \\
28th-layer &  0.8838 & \textbf{0.7134}     \\
24th-layer &  0.8801 & 0.6988     \\
20th-layer &  0.9063 & 0.6587     \\
middle-layer &  0.8123 & 0.650   \\
\midrule
BERT & 0.9403 & 0.5705\\
\bottomrule
\end{tabular}
\caption{Accuracy classifying truthfulness of sentences generated by the LLM itself. Unlike the data in Table \ref{tbl:generated}, where a threshold of 0.5 on the classifier output was used to classify sentences as true or false, in this table the results were obtained by estimating the optimal threshold from a held-out validation data set (30\% of the original test-set).}
\label{tbl:threshold}
\end{table}

Table \ref{tbl:threshold} presents the accuracy of all models when using the optimal threshold from the validation set. Clearly, SAPLMA performs better with a higher threshold. This somewhat confirms our assumption that the truth value of the sentences generated by the LLM is more subjective than those that appear in the true-false dataset. We note that also the BERT embeddings perform better with a higher threshold.
The use of a higher threshold can also be justified by the notion that it is better to delete or to mark as unsure a sentence that is actually true, than to promote false information.

Another interesting observation is that the 20th-layer no longer performs best for the statements generated by the LLM, but the 28th layer seems to perform best. This is somewhat perplexing and we do not have a good guess as to why this might be happening. Nevertheless, we stress that the differences between the accuracy levels of the 28th-layer and the others are statistically significant (using a two-tailed student t-test; $p<0.05$). In future work we will consider fusing multiple layers together, and using the intermediate activation values outputted by the LLM for all the words appearing in the statement (rather than using only the LLM's output for the final word).

We also ran the statements generated by the OPT 6.7b model on GPT-4 (March 23rd version), prompting it to determine whether each statement was true or false. Specifically, we provided the following prompt ``Copy the following statements and for each statement write true if it is true and false if it is false:'', and fed it $30$ statements at a time. It achieved an accuracy level of 84.4\%, and a Cohen's Kappa agreement level of 0.6883 with the true label. % (we note that each human annotator agreed with the majority over 97\% of the times).
%TODO: It would be interesting to see how well GPT-4 performs on the other topics. Probably much better.

\section{Discussion} % \& Limitations}
In this work we explicitly do not consider models that were trained or fine-tuned on data from the same topic of the test-set. This is particularly important for the sentences generated by the LLM, as training on a held-out set from them would allow the classifier to learn which \emph{type} of sentences generated by the LLM are generally true, and which are false. While this information may be useful in practice, and its usage is likely to yield much higher accuracy, it deviates from this paper's focus.

We note that the probability of the entire sentence (computed by multiplying the conditional probabilities of each word, given the previous words) cannot be directly translated to a truth value for the sentence, as many words are more common than others. Therefore, while sentence probabilities may be useful to determine which of two similar sentences is true, they cannot be used alone  for the general purpose of determining the truthfulness of a given sentence. 
%Amos: omitted to shorten the paper:
% For example, the false statement ``Vaduz is a city in Belgium'' has a probability of $6.44 \cdot 10^{-16}$. While this value is much lower than the corresponding true statement ``Vaduz is a city in Liechtenstein'', which has a probability of $2.31 \cdot 10^{-12}$, it is still much higher than the true statement ``Triesenberg is a city in Liechtenstein'', which has a probability of $3.21 \cdot 10^{-20}$. Therefore, any attempt to use these probabilities should, at minimum, normalize these probabilities by the frequency of the tokens in the data. Nevertheless, as we have discussed and shown, even when the LLM generates sentences using the most probable next token, it commonly generates false information.

In Table \ref{tbl:probablities} we compare the probability assigned by the LLM and the sigmoid output from SAPLMA on 14 statements, which do not appear in the true-false dataset. %, and were added explicitly to demonstrate the differences between the probabilities and SAPLMA 
We use the 28-layer, as it proved to perform best on the statements generated by the LLM, but we note that other layers provide very similar results on this set of statements. As depicted by the table, the probabilities provided by the LLM are highly susceptible to the syntax, i.e., the exact words and the statement's length.
The first two sets of examples in the table illustrate how sentence length highly affects the probabilities, but not SPLMA. In the following examples the false statements are not necessarily shorter than the true statements, yet the probabilities remain highly unreliable, while SPLMA generally succeeds in making accurate predictions.

The statement ``The Earth is flat'' as well as ``The Earth is flat like a pancake'' probably appear several times in the LLM's training data, therefore, it has a relatively high probability; however, SAPLMA is not baffled by it and is almost certain that both sentences are false. A basketeer is a rare word meaning a basketball player. It seems that the LLM is not familiar with the phrase and assigns it a low probability. However, while SAPLMA still classifies the statement ``Kevin Durant is basketeer'' as false, it's still much more confident that Kevin Durant is \emph{not} a baseball player, in contrast to the probabilities. Since the statement ``Kevin Duarnt is basketball player'' has a typo, its probability is extremely low, but SAPLMA still classifies the statement as true.

``Jennifer Aniston is a female person'' is an \emph{implicit truth}, a fact that is universally acknowledged without needing to be explicitly stated; thus, it is unlikely to be mentioned in the LLM's training data. Therefore, its probability is very low---much lower than ``Jennifer Aniston is not an actress''---despite having the same number of words. Nevertheless, SAPLMA classifies it correctly, albeit not with very high confidence.
%For example, since Kevin Durant is a basketball player, that exact sentence has a higher probability than the sentence stating that he is a baseball player. Similarly, SAPLMA is positive that the sentence ``Kevin Durant is a basketball player'' is true and that ``Kevin Durant is a baseball player'' is false. However, if instead of 
%has a high confidence that  quite positive that the sentence is true.  
%SAPLMA's values are much better aligned with the truth value. %We note that computing the probabilities requires access to the output of all previous 

%, and ``Carbon dioxide (CO2) is a colorless, odorless gas that consists of one carbon atom and two oxygen atoms,'' with a probablity of $2.22 \cdot 10^-16$. %``Neptune is the farthest planet in the solar system'' with a probability of $3.27 \cdot 10^-15$. %Carcassonne is a city in France

%Amos: removed to save space:
% We also attempt to record the probability of statements with ``It is true that ...'', and compare it to ``It is false that ...''. This way, each sentence is of the exact same length; however, this approach did not seem to perform well. For example, for the statement ``No dog has a tail,'' which is clearly false, the probability of ``It is true that no dog has a tail'' is $1.10 \cdot 10^{-16}$, which is much higher than ``It is false that no dog has a tail,'' which has a probability of only $3.34 \cdot 10^{-18}$. 
%All in all, w
While we show that the probabilities cannot be used alone to determine the veracity of a statement, they are not useless and do convey important information. Therefore, in future work we will consider providing SAPLMA with the probabilities of the generated words; however, this information may be redundant, especially if SAPLMA uses the intermediate activation values for all the words appearing in the statement. 

\begin{table}[htbp]
\centering
\resizebox{\columnwidth}{!}{
\begin{tabular}{l c c c}
\toprule
\textbf{Statement} & \textbf{Label} & \textbf{Probability} & \textbf{SAPLMA}\\
&&&(28th-layer) \\
\midrule
H2O is water, which is essential for humans & \textcolor{darkgreen}{True} & \textcolor{red}{6.64E-16} & \textcolor{darkgreen}{0.9032} \\
Humans don't need water & \textcolor{red}{False} & 
\textcolor{darkgreen}{2.65E-10} & \textcolor{red}{0.0282} \\
\midrule
The sun is hot, and it radiates its heat to Earth & \textcolor{darkgreen}{True} & \textcolor{red}{1.01E-17} & \textcolor{darkgreen}{0.9620} \\
The sun protects Earth from heat & \textcolor{red}{False} & \textcolor{darkgreen}{2.03E-14} & \textcolor{lightred}{0.3751} \\
\midrule
The Earth is flat & \textcolor{red}{False} & \textcolor{darkgreen}{5.27E-07} & \textcolor{red}{0.0342} \\
The world is round and rotates & \textcolor{darkgreen}{True} & \textcolor{red}{2.96E-11} & \textcolor{lightgreen}{0.6191} \\
The Earth is flat like a pancake & \textcolor{red}{False} & \textcolor{redgreen}{3.88E-10} & \textcolor{red}{0.0097} \\
\midrule
Kevin Durant is a basketball player& \textcolor{darkgreen}{True} & \textcolor{darkgreen}{2.89E-10} & \textcolor{darkgreen}{0.9883} \\
Kevin Durant is a baseball player & \textcolor{red}{False} & \textcolor{redgreen}{4.56E-12} & \textcolor{red}{0.0001} \\
Kevin Durant is a basketeer & \textcolor{darkgreen}{True} & \textcolor{red}{5.78E-16} & \textcolor{red}{0.0469} \\
Kevin Duarnt is a basketball player	& \textcolor{darkgreen}{True} & \textcolor{darkred}{1.52E-21} & \textcolor{lightgreen}{0.7105}\\
%Kevin Duarnt is a basketeer & True & 3.76E-26 & 0.0484 \\
\midrule
Jennifer Aniston is an actress & \textcolor{darkgreen}{True} & \textcolor{darkgreen}{1.88E-10} & \textcolor{darkgreen}{0.9985}\\
Jennifer Aniston is not an actress & \textcolor{red}{False} & \textcolor{redgreen}{1.14E-11} & \textcolor{red}{0.0831}\\
Jennifer Aniston is a female person & \textcolor{darkgreen}{True} & \textcolor{red}{2.78E-14} & \textcolor{lightgreen}{0.6433}\\
\midrule
Harry Potter is real & \textcolor{red}{False} & \textcolor{darkgreen}{9.46E-09} & \textcolor{red}{0.0016} \\
Harry Potter is fictional & \textcolor{darkgreen}{True} & \textcolor{redgreen}{1.53E-09} & \textcolor{darkgreen}{0.9256} \\
Harry Potter is an imaginary figure & \textcolor{darkgreen}{True} & \textcolor{red}{6.31E-14} & \textcolor{darkgreen}{0.8354} \\
\bottomrule
\end{tabular}
}
\caption{Comparison of the probability assigned by the LLM and the sigmoid output from SAPLMA's 28th layer (using OPT-6.7b), color-coded for clarity. SAPLMA's values are much better aligned with the truth value.}
\label{tbl:probablities}
\end{table}

%Vaduz is a city in Belgium.---Sentence Probablity: 6.440917853888796e-16
%Vaduz is a city in Liechtenstein.---Sentence Probablity: 2.3152273916544473e-12
%Carbon dioxide (CO2) is a colorless, odorless gas that consists of one carbon atom and two oxygen atoms.---Sentence Probablity: 2.225848066192215e-16
%Triesenberg is a city in France.---Sentence Probablity: 1.0396081013444022e-20
%Triesenberg is a city in Liechtenstein.---Sentence Probablity: 3.2108583328354813e-20
%Vaduz is a city in France.---Sentence Probablity: 6.88500540810185e-18
%Triesenberg is a city in Belgium.---Sentence Probablity: 1.05496479874502e-19

\section{Conclusions \& Future Work}
In this paper, we tackle a fundamental problem associated with LLMs, i.e., the generation of incorrect and false information. We have introduced SAPLMA, a method that leverages the hidden layer activations of an LLM to predict the truthfulness of generated statements. We demonstrated that SAPLMA outperforms few-shot prompting in detecting whether a statement is true or false, achieving accuracy levels between 60\% to 80\% on specific topics when using OPT-6.7b, and between 70\% to 90\% when using LLAMA2-7b. This is a significant improvement over the maximum 56\% accuracy level achieved by few-shot prompting (for OPT-6.7b).

Our findings suggest that LLMs possess an internal representation of statement accuracy, which can be harnessed by SAPLMA to filter out incorrect information before it reaches the user, and furthermore that this representation of accuracy is very different from the probability assigned to the sentence by the LLM. Using SAPLMA as a supplement to LLMs may increase human trust in the generated responses and mitigate the risks associated with the dissemination of false information. Furthermore, we have released the true-false dataset and proposed a method for generating such data. This dataset, along with our methodology, provides valuable resources for future research in improving LLMs' abilities to generate accurate and reliable information.

In future work we intend to apply our method to larger LLMs, and run experiments with humans, such that a control group will interact with an unfiltered LLM, and the experimental group will interact with a system that augments SAPLMA and an LLM. We hope to demonstrate that humans trust and better understand the limitations of a system that is able to review itself and mark statements that it is unsure about. We also intend to study how the activations develop over time as additional words are generated, and consider multilingual input. %, so that SAPLMA can be applied to detect inaccurate information within a passage generated by an LLM.

\section{Limitations}
This paper focuses on detecting whether a statement is true or false. However, in practice, it may be more beneficial to detect if the LLM is positive that a statement is correct or if it is unsure. The most simple adjustment to the proposed method in this paper is to lift the required threshold for classifying a statement as true above $0.5$; however, the exact value would require some form of calibration of the model \cite{bella2010calibration}. Another option is to use multiple classifiers and to require all (or a vast majority of) classifiers to output ``true'', for the statement to be considered true. Alternatively, dropout layers can be used for the same goal \cite{chen2021adaptive}. The overall system can benefit from multiple outcomes, such that if the LLM is not positive whether the statement is true or false, it can be marked for the user to treat with caution. However, if a statement is classified as false, the LLM can delete it and generate a different statement instead. To avoid regenerating the same statement again, the probability of sampling the words that appear in the current statement should be adjusted downward.

Our work was only tested in English. However, we believe that a multilingual LLM can be trained on one language and applied on statements in another language. We will test this hypothesis in future work. 
  
In our work, we collected the activation values when each sentence was generated separately. However, in practice, in an LLM generating longer responses the activation values develop over time, so they may process both correct and incorrect information. Therefore, the activation values would need to be decoupled so that they can be tested whether the most recent statement was true or false. One approach might be to subtract the value of the activations obtained after the previous statement from the current activation values (a discrete derivative). Clearly, training must be performed using the same approach. %This might 

\section{Ethical Impact}
One of the primary ethical concerns of LLMs is the generation of false information; yet, we believe that SAPLMA could potentially reduce this issue. On the other hand, it is important to acknowledge that certain ethical issues, such as bias, may persist, potentially being transferred from the LLM to SAPLMA. Specifically, if the LLM exhibits bias towards certain ethnic groups, SAPLMA may likewise classify statements as true or false based on these inherited biases from the original LLM. Nevertheless, it may be possible to adapt the  approach presented in this paper to bias mitigation.

\section{Acknowledgments}
This work was supported, in part, by the Ministry of Science and Technology, Israel, and by the Israel Innovation Authority.

\bibliographystyle{acl_natbib}
\bibliography{citations}

\end{document}